\documentclass[11pt]{article}
\usepackage{coling2020}
\usepackage{times}
\usepackage{url}
\usepackage{latexsym}

\usepackage{coling-nert}
\usepackage{microtype}

\colingfinalcopy % Uncomment this line for the final submission

\title{A Human Evaluation of AMR-to-English Generation Systems}

\author{Emma Manning \quad Shira Wein \quad Nathan Schneider \\ 
  Georgetown University \\
  {\{\emldisplay{esm76@georgetown.edu}{esm76}, \emldisplay{sw1158@georgetown.edu}{sw1158},
  \emldisplay{nathan.schneider@georgetown.edu}{nathan.schneider}\}\texttt{@georgetown.edu} }}

\date{}

\begin{document}
\maketitle
\begin{abstract}
Most current state-of-the art systems for generating English text from Abstract Meaning Representation (AMR) have been evaluated only using automated metrics, such as BLEU, which are known to be problematic for natural language generation. In this work, we present the results of a new human evaluation which collects fluency and adequacy scores, as well as categorization of error types, for several recent AMR generation systems. We discuss the relative quality of these systems and how our results compare to those of automatic metrics, finding that while the metrics are mostly successful in ranking systems overall, collecting human judgments allows for more nuanced comparisons. We also analyze common errors made by these systems.
\end{abstract}

\section{Introduction}
\label{sec:intro}

\blfootnote{
    \hspace{-0.65cm}  % space normally used by the marker
    This work is licensed under a Creative Commons Attribution 4.0 International License.\\
    License details:
    \url{http://creativecommons.org/licenses/by/4.0/}.
}

Abstract Meaning Representation, or AMR \citep{banarescu-etal-2013-abstract}, is a representation of the meaning of a sentence as a rooted, labeled, directed acyclic graph. For example,\\

\noindent\texttt{
(l / label-01\\
 \indent     :ARG0 (c / country :wiki "Georgia\_(country)"
           :name (n / name :op1 "Georgia"))\\
 \indent     :ARG1 (s / support-01\\
   \indent\indent         :ARG0 (c2 / country :wiki "Russia"
             :name (n2 / name :op1 "Russia")))\\
\indent      :ARG2 (a / act-02
        :mod (a2 / annex-01)))}\\ 
 
\noindent represents the sentence ``Georgia labeled Russia's support an act of annexation.'' AMR does not represent some morphological and syntactic details such as tense, number, definiteness, and word order; thus, this same AMR could also represent alternate phrasings such as ``Russia's support is being labeled an act of annexation by Georgia.''
 
 AMR generation is the task of generating a sentence in natural language (in this case, English) from an AMR graph. This has applications to a range of NLP tasks, including summarization \citep{liao-etal-2018-abstract} and machine translation \citep{song-etal-2019-semantic}.
 Like other Natural Language Generation (NLG) tasks, this is difficult to evaluate due to the range of possible valid sentences corresponding to any single AMR.
 
Currently, AMR generation systems are often evaluated only with automatic metrics that compare a generated sentence to a single human-authored reference; for AMR, this is the sentence from which the AMR graph was created. However, there is evidence that these metrics may not be a good representation of human judgments for AMR generation \citep{may-priyadarshi-2017-semeval} and NLG in general (see \cref{ssec:metrics}).

Thus, in this work, we present a new human evaluation of several recent AMR generation systems, most of which had not previously been manually evaluated. Our methodology (\cref{sec:methodology}) differs in several ways from previous evaluations of AMR generation, including separate direct assessment of fluency and adequacy; and asking annotators to evaluate sentences without comparison to a reference, in order to avoid biasing them toward a particular wording. We analyze (\cref{sec:analysis}) what our results show about the relative quality of the systems  and how this compares to their scores from automatic metrics, finding that these metrics are mostly accurate in ranking systems, but that collecting separate judgments for fluency, adequacy, and error types allows us to characterize the relative strengths and weaknesses of each system in more detail. Finally, we discuss common errors among sentences which received low scores from annotators, identifying issues for future researchers to address including hallucination, anonymization, and repetition.

\section{Background}

In \cref{ssec:metrics} we discuss previous work on evaluation of AMR generation and related NLG tasks, both with automatic metrics and human evaluation. In \cref{ssec:systems} we survey recent work in AMR generation, including describing the systems which we evaluate.

\subsection{Evaluation of AMR Generation}

\paragraph{Automatic Metrics:}
\label{ssec:metrics}
The vast majority of AMR generation papers measure their performance only with automatic metrics. The most common of these metrics is BLEU \citep{papineni-etal-2002-bleu}, which is typically used to determine the state of the art. However, it is unclear whether BLEU is a reliable metric to compare AMR generation systems: \citet{may-priyadarshi-2017-semeval} found that BLEU disagreed with human judgments on the ranking of five AMR generation systems, including disagreeing on which system was the best. Concerns have also been raised about the suitability of BLEU for NLG in general; for example, \citet{reiter-2018-structured} found that BLEU has generally poor correlations with human judgments for NLG. \citet{novikova-etal-2017-need} compared many metrics to human judgments on NLG from meaning representations and concluded that use of reference-based metrics relies on an invalid assumption that references are correct and complete enough to be used as a gold standard.

Some recent AMR generation papers have reported other automatic metrics alongside BLEU. Many have reported METEOR \cite{banerjee-lavie-2005-meteor}, and a few have included TER \cite{Snover2006} and, more recently, CHRF++ \cite{popovic-2017-chrf} and BERTScore \citep{zhang2019bertscore}. However, it is unclear how accurately any of these metrics capture the relative performance of AMR generation systems.

\paragraph{Human Evaluation:}

Prior to this work, the only human evaluation comparing several AMR generation systems was the SemEval-2017 AMR shared task, which used a ranking-based evaluation of five systems \citep{may-priyadarshi-2017-semeval}. All of these systems perform far below the current state-of-the-art, making a new evaluation necessary.

While most AMR generation papers have reported no human evaluation of their systems, a few have conducted smaller-scale evaluations. \Citet{ribeiro-etal-2019-enhancing} conducted a Mechanical Turk evaluation to compare their best graph encoder model with a sequence-to-sequence baseline, finding that their model performs better on both meaning similarity between the generated sentence and the gold reference, and readability of the generated sentence. 

\Citet{mager-etal-2020-gpt} carry out a human evaluation of overall quality, comparing their GPT-2-based system to three others \citep{guo-etal-2019-densely, ribeiro-etal-2019-enhancing, zhu-etal-2019-modeling}, all of which are also evaluated in our experiment. For the three systems included in both our evaluation and theirs, the relative results are comparable; \citet{mager-etal-2020-gpt} find their own system to be better than all three.

\Citet{Lapalme2019} also conducted a small human evaluation in which annotators chose the best output out of three options: their own system,  ISI  \citep{pourdamghani-etal-2016-generating}, and JAMR \citep{flanigan-etal-2016-generation}. They find that their rule-based system is on par with ISI and much better than JAMR, despite having a much lower BLEU.

Beyond AMR generation, other NLG tasks are also often evaluated only with automatic metrics; for example, \citet{gkatzia-mahamood-2015-snapshot} found that 38.2\% of NLG papers overall, and 68\% of those published in ACL venues, used automatic metrics. However, as discussed above, many studies have found that these metrics are not a reliable proxy for human judgments. %and argued for the use of manual evaluation. 
One example of the use of human evaluation is the Conference on Machine Translation (WMT), which runs an annual evaluation of machine translation systems \citep[e.g.][]{barrault-19}.

\subsection{Recent Advances in AMR Generation}
\label{ssec:systems}
Shortly after the 2017 shared task, \citet{konstas-etal-2017-neural} made significant advances to the field with a neural sequence-to-sequence approach, mitigating the limitations of the small amount of AMR-annotated data by augmenting training data with a jointly-trained parser. 

Later work \citep{song-etal-2018-graph} builds on this approach but uses a graph-to-sequence model to preserve information from the structure of the AMR. Several recent papers have explored variations on a graph-to-sequence approach: improvements in encoding reentrancies and long-range dependencies \citep{damonte-cohen-2019-structural}, a dual graph encoder that captures top-down and bottom-up representations of graph structure \citep{ribeiro-etal-2019-enhancing}, and a densely-connected graph convolutional network \citep{guo-etal-2019-densely}.

Recent sequence-to-sequence approaches include using structure-aware self-attention to capture relations between concepts within a sequence-to-sequence transformer model \citep{zhu-etal-2019-modeling}, generating syntactic constituency trees as an intermediate step before generating surface structure \citep{cao-clark-2019-factorising}, and fine-tuning GPT-2 on AMR \citep{mager-etal-2020-gpt}.

While neural approaches have achieved state-of-the-art BLEU scores, a few recent works have instead approached AMR generation through more rule-based methods. \citet{manning-2019-partially} constrains their system with rules, supplemented by simple statistical models, to avoid certain types of errors, such as hallucinations, that are possible in neural systems. \citet{Lapalme2019} create a fully rule-based generation system to help humans check their AMR annotations. 

\section{Methodology}
\label{sec:methodology}

We conduct a human evaluation of several AMR generation systems. \Cref{ssec:design} discusses the general survey design, while \cref{ssec:pilot_methods} discusses details of the pilot survey, which validates the methodology by applying it to data from the SemEval evaluation, and \cref{ssec:main_methods} discusses the evaluation of more recent systems.

\begin{figure}[htb]
\centering
\includegraphics[width=0.45\textwidth]{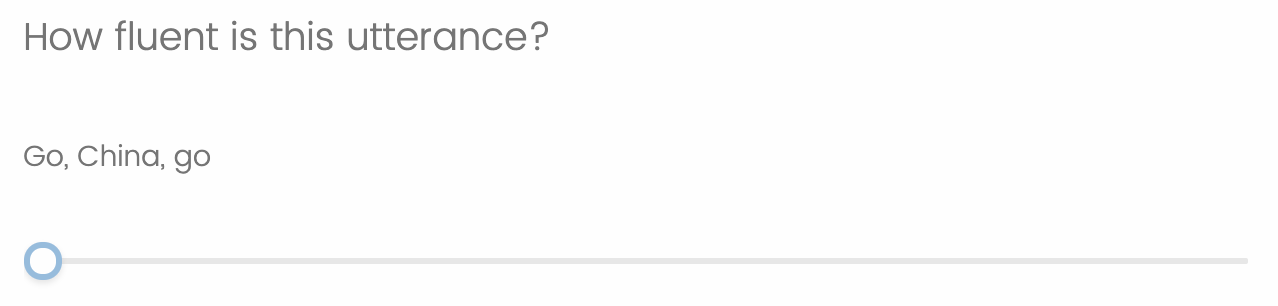}
\caption{\label{fig:screenshot-fluency}Screenshot from the fluency section of the survey.}
\end{figure}
\begin{figure}[htb]
\centering
\includegraphics[width=0.45\textwidth]{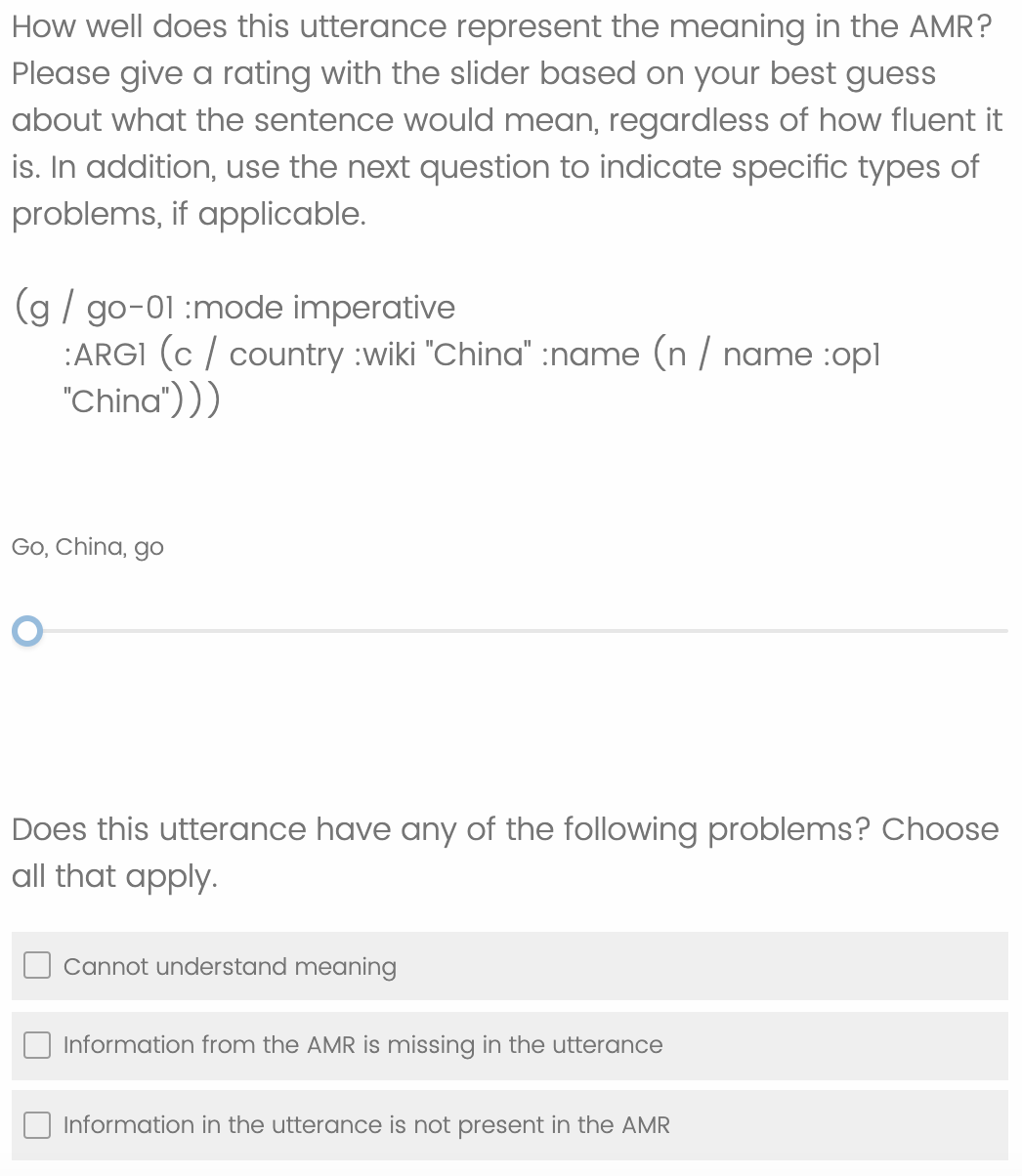}
\caption{\label{fig:screenshot-adequacy}Screenshot from the adequacy section of the survey.}
\end{figure}

\subsection{Survey Design}\label{ssec:design}
\Cref{fig:screenshot-fluency,fig:screenshot-adequacy} show examples of the survey interface for one sentence.

\paragraph{Scalar Scores:}

The SemEval-2017 evaluation of AMR generation elicited judgments in the form of relative rankings of output from three systems at a time \citep{may-priyadarshi-2017-semeval}. However, recent work in evaluation of machine translation \citep{bojar-etal-2016-findings} has found that direct assessment is a preferable method to collect judgments, partly because it evaluates absolute quality of translations. We use a similar direct assessment method, providing annotators with a slider which represents scores from 0 to 100, although annotators are not shown numbers. Unlike recent WMT evaluations, we collect separate scalar scores for \textbf{fluency} and \textbf{adequacy}. This has been common practice in many evaluations of NLG and MT; for example, \citet{Gatt2010a} also use separate direct assessment sliders for these two dimensions for NLG.

\paragraph{Referenceless Design:}
Many human evaluations of NLG and MT, including the SemEval evaluation for AMR, provide a reference for the annotator to compare to the system output. However, since AMR is underspecified with respect to many aspects of phrasing including tense, number, word order, and definiteness, comparison to a single reference risks biasing annotators toward the specific phrasing used in the reference. Thus, each survey given to annotators consists of two sections: in the first half, annotators judged fluency, and saw only the output sentences; in the second, they judged the same sentences on adequacy, and were shown the AMR from which the sentence was generated, allowing them to compare the meanings. This design required that our annotators be familiar with the AMR scheme to identify mismatches in the concepts and relations expressed in the sentences.

\paragraph{Adequacy Error Types:}
In addition to numeric scores, under each adequacy slider are three checkboxes where annotators can indicate whether certain types of adequacy errors apply:
\begin{itemize}
    \item That they cannot understand the meaning of the utterance (i.e.~it is disfluent enough to be incomprehensible, making it difficult to meaningfully judge adequacy)
    \item That information in the AMR is missing from the utterance
    \item That information not present in the AMR is added in the utterance
\end{itemize}
These options allow for a more nuanced analysis of the types of mistakes made by different systems than numerical scores alone would provide.

\paragraph{Survey Structure:}
Instructions for judging fluency are provided at the beginning of the survey, and instructions for adequacy are shown before the start of the adequacy portion. For fluency, annotators are asked to ``indicate how well each one represents fluent English, like you might expect a person who is a native speaker of English to use,'' and told that ``some of these may be sentence fragments rather than complete sentences, but can still be considered fluent utterances.'' For adequacy, they are instructed to ``determine how accurately the sentence expresses the meaning in the AMR.'' The full text of these instructions, which also includes examples, is provided in the supplementary material. 

Each page of the survey includes each system's output for a given sentence, presented in a random order. The reference is also included as a sentence to judge, but is not distinguished from the system outputs.

\subsection{Pilot Evaluation}\label{ssec:pilot_methods}

Before collecting the full dataset of human judgments for AMR generation, we completed a smaller pilot experiment to test the validity and practicality of the methodology. This pilot used the data and systems included in the SemEval-2017 shared task  \citep{may-priyadarshi-2017-semeval}. A random subset of 25 out of the 1293 sentences in the dataset were used. All were annotated by three annotators, each of whom was a native speaker of English and a linguist with experience with AMR.

We tweaked the design of the later survey based on feedback from the pilot annotators. In particular, the surveys were shortened (annotators completed two batches of 10 sentences each, instead of one with 25); more thorough instructions were given, with examples; and wording was changed from ``sentence'' to ``utterance'' to reflect that some are not full sentences in a grammatical sense.

\subsection{Main Evaluation}\label{ssec:main_methods}
The main evaluation was larger than the pilot, and evaluated more recent systems, most of which are of a markedly higher quality than those in the pilot. We contacted the authors of several recent papers on AMR-to-English generation to obtain their system's output for use in the evaluation, and included all systems for which we obtained usable data in time to begin evaluation: \citet{konstas-etal-2017-neural}, \citet{guo-etal-2019-densely}, \citet{manning-2019-partially}, \citet{ribeiro-etal-2019-enhancing}, and \citet{zhu-etal-2019-modeling}. These systems are all described in \cref{ssec:systems}. 

The Konstas and Zhu systems both approach AMR generation as a sequence-to-sequence task: Konstas is pretrained on `silver' data from a jointly-trained AMR parser to mitigate the limitations of the small amount of gold AMR data, while Zhu uses a transformer-based model. Guo and Ribeiro both use graph-to-sequence models: Guo uses densely-connected graph convolutional networks to model the graph structure, while Ribeiro uses a dual representation to capture both bottom-up and top-down perspectives on the graph. Finally, Manning is the only non-neural system in this evaluation; it uses handwritten rules to generate possible realizations, augmented with simple statistical modules to choose between candidates.

\paragraph{Data:}
We use the test set from the LDC2017T10 AMR dataset \citep{amr2017}. However, some of the system output we obtained was generated from the earlier 2015 version of this data (LDC2015E86); these test sets contain the same sentences, with some updates to the AMRs, so we decided to only include AMRs at the intersection of these datasets in our evaluation.

Additionally, we chose to exclude AMRs whose root relation was \texttt{multisentence}, which indicates that the portion of text officially segmented as one sentence includes what AMR annotators analyzed as two or more sentences.
These were excluded because they are often very long and pilot annotators found they could be very difficult to read and evaluate, and because unlike other AMR relations, \texttt{multisentence} does not represent a semantic relationship between elements of meaning.

A total of 335 sentences were excluded from consideration due to differences in their AMRs between the different versions of the data, and 71 for being multi-sentence. Accounting for overlap between the excluded sets, 998 out of 1371 total sentences in the test set were considered eligible for our evaluation. A random sample of 100 of these were used in the survey.

\paragraph{Annotation:}
A total of nine annotators participated in this evaluation, including the three who participated in the pilot. All had prior training in AMR annotation, mostly from taking a semester-long course focused on AMR and other meaning representations. Six of the annotators were native English speakers; all were either current or former PhD students or professors at our university.

Each person annotated two different batches of 10 sentences each, except for one annotator who did four batches. The result was that each set of sentences was double-annotated, allowing us to quantify inter-annotator agreement. Additionally, batches were assigned such that each annotator overlapped with at least two other annotators.

\section{Analysis}
\label{sec:analysis}

\subsection{Survey Reliability}

\paragraph{Pilot:} 
The only previous human evaluation of several AMR-to-English generation systems was in the SemEval-2017 task discussed above. Since our survey had several differences from this previous evaluation, it was possible that the methodological differences could lead to substantial differences in judgments on the same data. Thus, before conducting the main survey, we validated our methodology by comparing the results of the pilot survey to that of the SemEval-2017 evaluation.

This is the first evaluation of AMR generation to collect separate judgments for fluency and adequacy. We hypothesized that this would provide a finer-grained characterization of system behavior, and that annotators would be able to distinguish these two scales, though they are related (incomprehensible sentences necessarily have low fluency as well as accuracy, while references and high-quality output have near-perfect fluency and adequacy). Indeed, we find a Spearman's rank correlation of 0.68 between fluency and adequacy ratings in the pilot, indicating that while they are related, annotators were largely able to evaluate these two dimensions separately. 

The average fluency scores from our evaluation match the ranking of systems found in \citet{may-priyadarshi-2017-semeval}. Average adequacy scores are the same except that the third- and fourth-place systems switch places. This suggests that our methodology is reliable for ranking systems, and that separating judgments for fluency and adequacy allows for a more nuanced view of relative system performance than overall quality judgments.

Finally, we calculate inter-annotator agreement (IAA) to measure how consistently annotators could make these judgments. We measure IAA for the numeric fluency and adequacy scores with Spearman's correlation, and for each adequacy error type with Cohen's Kappa.

We find an average pairwise IAA of 0.78 for fluency and 0.67 for adequacy. For error types, we get lower agreement: average pairwise Kappa scores are 0.44 for incomprehensibility, 0.53 for missing information, and 0.28 for added information. This indicates that guidelines on when to annotate these error types were not made clear enough for annotators to apply them consistently; future studies using this methodology should clarify these guidelines for more reliable results.

\begin{table}[htb]
\centering\small
\begin{tabular}{|c|rr|rrr|}
\hline \textbf{Pair \#} & \textbf{F} & \textbf{A} & \textbf{INC} & \textbf{MI} & \textbf{AI} \\ \hline
0 & 0.49 & 0.71 & 0.41 & 0.11 & 0.19 \\
1 & 0.83 & 0.85 & 0.13 & 0.30 & 0.09 \\
2 & 0.63 & 0.79 & 0.57 & 0.24 & 0.64 \\
3 & 0.25 & 0.51 & 0.46 & 0.47 & -0.02 \\
4 & 0.52 & 0.49 & 0.28 & 0.46 & 0.47 \\
5 & 0.16 & 0.54 & 0.12 & 0.36 & 0.47 \\
6 & 0.44 & 0.53 & 0.24 & 0.57 & 0.31 \\
7 & 0.41 & 0.54 & 0.48 & 0.67 & 0.37 \\
8 & 0.82 & 0.79 & 0.73 & 0.71 & 0.52 \\
9 & 0.60 & 0.74 & 0.30 & 0.52 & 0.44 \\
\hline
\textbf{AVG} & \textbf{0.51} & \textbf{0.65} & \textbf{0.37} & \textbf{0.44} & \textbf{0.35}\\
\hline
\end{tabular}
\caption{\label{tab:iaa} Inter-annotator agreement scores for each annotator pair. For numeric ratings of Fluency (F) and Adequacy (A), we use Spearman's Rho; for binary categorical ratings of Incomprehensibility (INC), Missing Information (MI), and Added Information (AI), we use Cohen's Kappa.}
\end{table}

\paragraph{Main Survey:}
On this survey we find an overall Spearman's correlation of 0.58 between fluency and adequacy, indicating that annotators were able to evaluate these two dimensions separately. This correlation is lower than in the pilot, which may be due to clearer instructions given to annotators on what is meant by ``fluency'' and ``adequacy'', or because the two dimensions are easier to separate when fewer sentences are of very low quality. 

Since each set of 10 AMRs (60 judgments of each type per annotator) was double-annotated by a different pair of annotators, we evaluate IAA separately for each pair. Results are shown in \cref{tab:iaa}. Agreement scores vary considerably, but indicate moderate agreement overall. We find that IAA for fluency is moderate to high for most annotator pairs, with two exceptions. IAA is higher for adequacy than for fluency in 8 out of 10 cases, and reflects at least moderate agreement in all cases.

For adequacy error types, IAA scores vary greatly and many are low. This indicates that guidelines given to annotators may not have been clear enough. For example, it was expected that annotators would infer, based on their knowledge that AMR does not specify tense, that sentences should not be considered wrong for having any particular tense; however, we learned after the evaluation that at least one annotator marked some cases of non-present tense in sentences as added information. 

\begin{figure*}[htb]
    \centering
    \begin{subfigure}[b]{0.44\textwidth}
         \includegraphics[width=\textwidth]{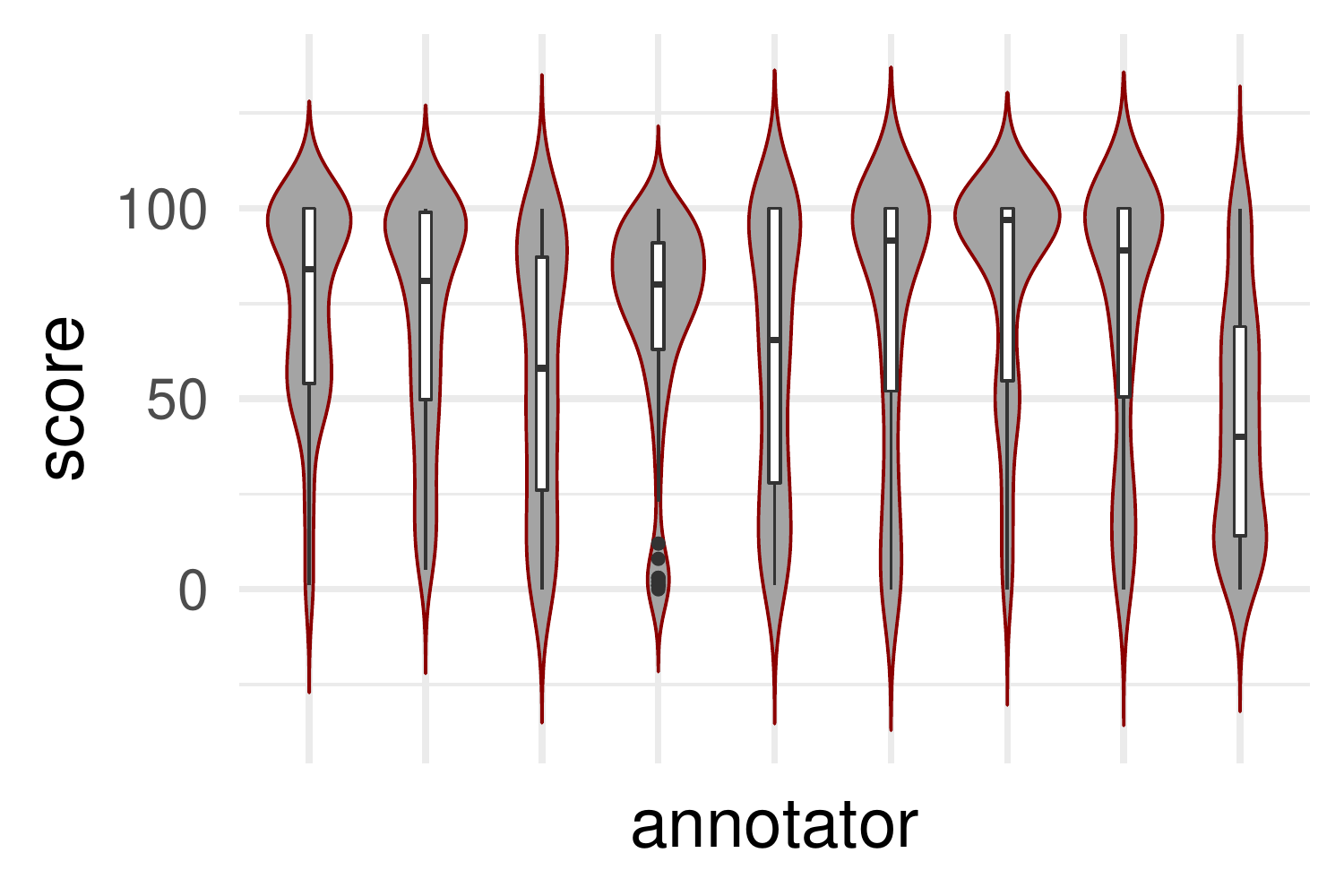}
        \caption{\label{fig:annotator-fluency}Fluency by annotator}
    \end{subfigure}
    \hspace*{3em}
    \begin{subfigure}[b]{0.44\textwidth}
        \includegraphics[width=\textwidth]{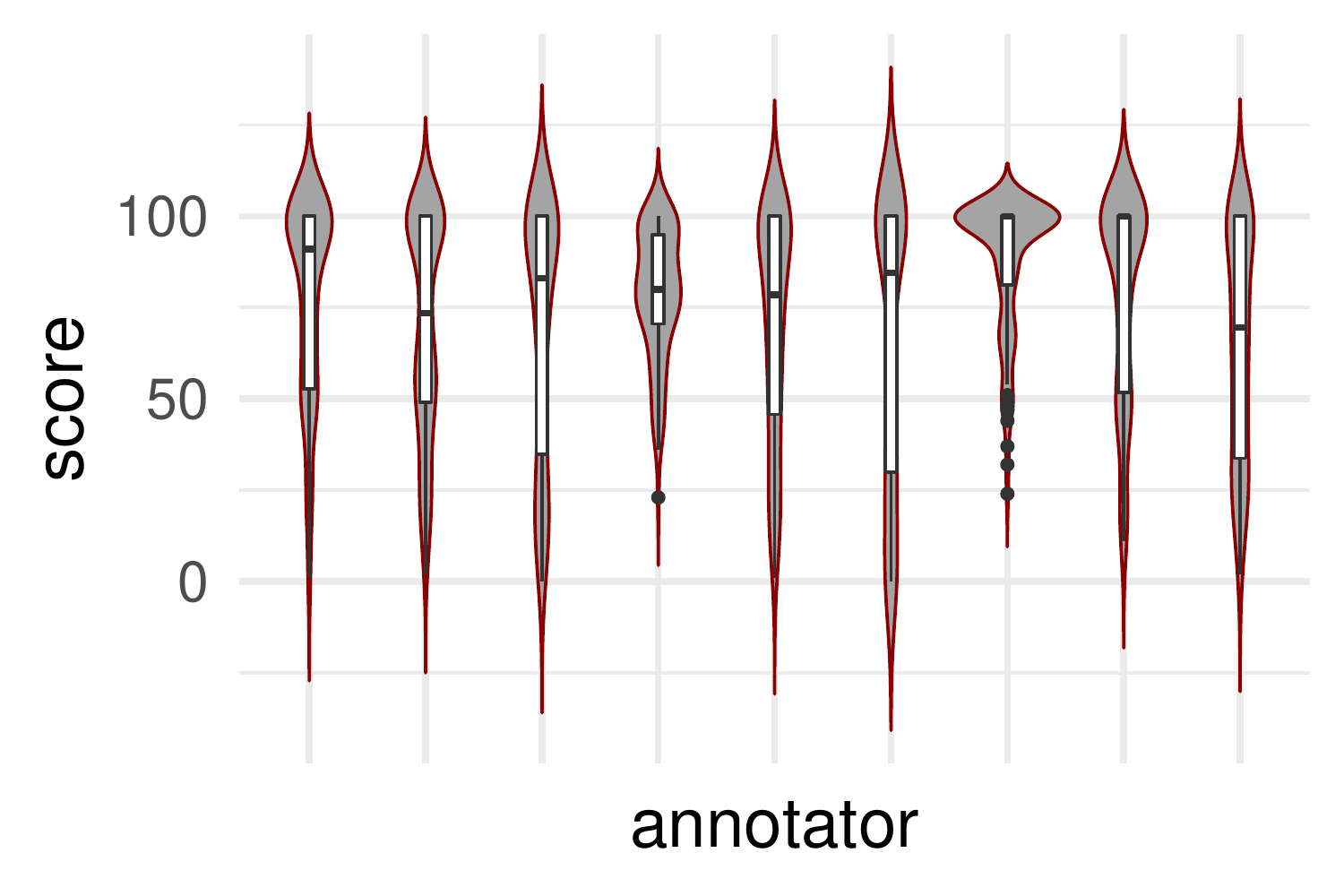}
        \caption{\label{fig:annotator-adequacy}Adequacy by annotator}
    \end{subfigure}
    \caption{Violin plots of ranges of human judgments for each annotator.}\label{fig:annotator-violins}
\end{figure*}

\Cref{fig:annotator-violins} gives each annotator's distribution of ratings, showing that different individuals distributed their judgments over the available \mbox{0--100} scale in different ways. Since each annotator judged each system the same number of times, this is not a problem for comparison of systems. However, when identifying low-scoring sentences (\cref{ssec:low-adequacy,ssec:low-fluency}), we normalize by annotator to account for these differences.

\subsection{Quality of Systems}

\begin{table*}
\begin{minipage}{.6\textwidth}
%\begin{table}
\centering
\small
\begin{tabular}{|l|r@{~~}>{\smaller}rr@{~~}>{\smaller}r|rrr|}
\hline \textbf{System} & \multicolumn{2}{c}{\textbf{F}$_\uparrow$} & \multicolumn{2}{c|}{\textbf{A}$_\uparrow$} & \textbf{INC}$_\downarrow$ & \textbf{MI}$_\downarrow$ & \textbf{AI}$_\downarrow$ \\ \hline
Konstas & \textbf{78.14} & 1 & \textbf{81.46} & 1 & \textbf{10.0} & 34.5 & 12.0 \\
Zhu & 71.61 & 2 & 74.13 & 2 & 15.5 & 36.0 & 25.5 \\
Ribeiro & 67.05 & 3 & 64.37 & 4 & 19.5 & 47.0 & 31.5 \\
Guo & 62.13 & 4 & 68.52 & 3 & 22.0 & 41.0 & 21.5 \\
Manning & 36.89 & 5 & 54.10 & 5 & 57.5 & \textbf{17.5} & \textbf{\hphantom{0}9.0} \\
\hline
Reference &87.56 & & 93.68 & & \hphantom{0}5.0 & \hphantom{0}4.5 & 10.0 \\
\hline
\end{tabular}
\caption{\label{tab:avg-scores} For each system, average fluency and adequacy scores and percentage where each adequacy error type was selected. Scores for the reference sentences are included for comparison.}
\end{minipage}
\quad
\begin{minipage}{.35\textwidth}
\centering\small
\begin{tabular}{|l|rr|}
\hline \textbf{System} & \textbf{\# low F} & \textbf{\# low A} \\ \hline
Konstas & 5 & 9\\
Zhu &  9 & 16\\
Ribeiro & 21 & 34\\
Guo & 21 & 28\\
Manning & 60 & 51 \\
Reference & 0 & 1 \\
\hline
Total & 116 & 139 \\
\hline
\end{tabular}
\caption{\label{tab:low-scores} Of 100~sentences, number with low fluency or adequacy (bottom 1$/$3 of both annotators' scores).}
%\end{table}
\end{minipage}
\end{table*}

\Cref{tab:avg-scores} shows the average score given for each system for fluency and adequacy, as well as how often each was marked as having each adequacy error type. We find that on both fluency and adequacy scores, Konstas performs best, followed by Zhu, and Manning performs the worst. Guo and Ribeiro are in between and within 5 points of each other on each measure, with Ribeiro performing better on fluency and Guo on adequacy.

Unsurprisingly, the lower a system's average fluency score, the more often sentences were marked as incomprehensible.

The Missing Information and Added Information labels support the suggestion of \citet{manning-2019-partially} that although their system performs worse than others by most measures, its constraints make it less likely than machine-learning-based systems to omit or hallucinate information. Konstas's system performs the next-best by both of these measures; in particular, it rarely adds information not present in the AMR. Ribeiro's system is most prone to errors of these types, omitting information in nearly half of sentences and hallucinating it in nearly a third. Overall, the results from these questions indicate that neural AMR generation systems are prone to omit or hallucinate concepts from the AMR with concerning frequency.

\begin{figure*}[htb]
    \centering
    \begin{subfigure}[b]{0.4\textwidth}
        \includegraphics[width=\textwidth,height=5cm]{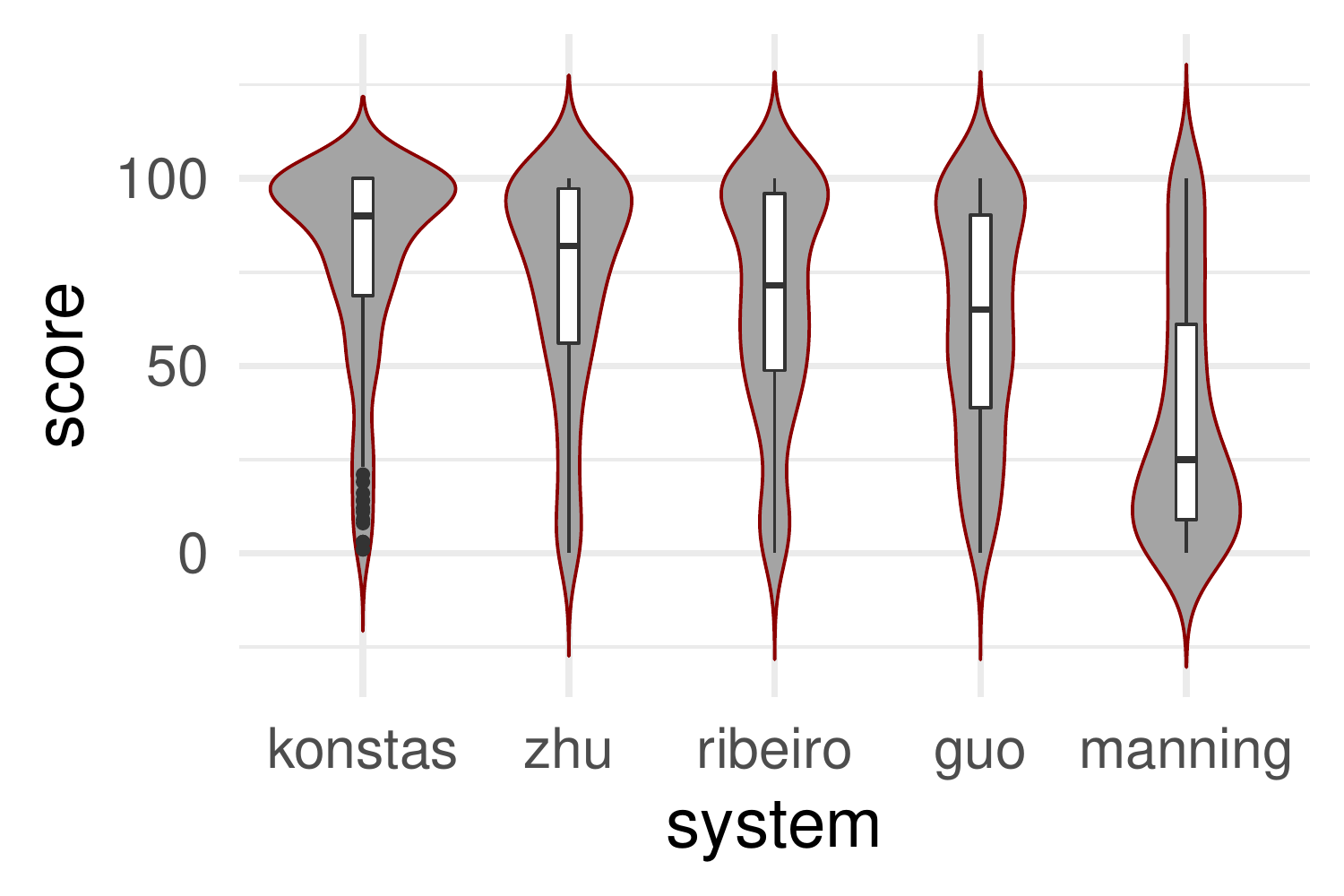}
        \caption{\label{fig:system-fluency}Fluency by system}
    \end{subfigure}
    \hspace*{4em}
    \begin{subfigure}[b]{0.4\textwidth}
        \includegraphics[width=\textwidth,height=5cm]{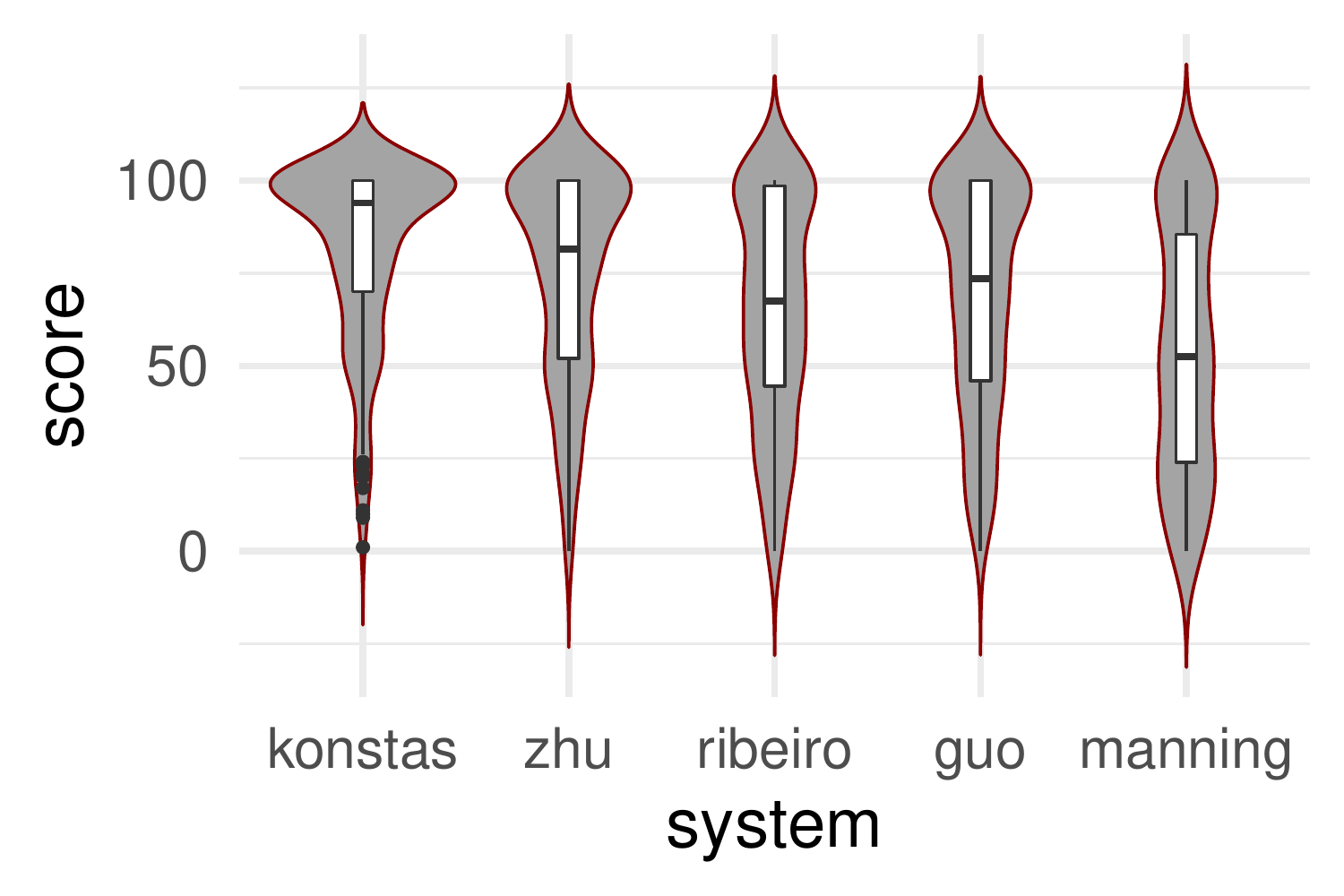}
        \caption{\label{fig:system-adequacy}Adequacy by system}
    \end{subfigure}
    \caption{Violin plots of human judgments for each system. \label{fig:violin_system}}
\end{figure*}

\Cref{fig:system-fluency,fig:system-adequacy} show the distributions of scores each system received for fluency and adequacy, respectively.\footnote{Reference scores are omitted from these figures because the high concentration of perfect scores obscured the details of other systems.} These show that Konstas is skewed toward very high scores, and that Manning skews toward low scores especially for fluency.

% \begin{figure*}
%     \centering
%     %\setlength{\belowcaptionskip}{0cm}
%     \begin{subfigure}[b]{0.4\textwidth}
%         \includegraphics[width=\textwidth]{bleu_fluency_new.pdf}
%         \caption{\label{fig:bleu-fluency}Comparison of BLEU scores to average fluency scores from human evaluation.}
%     \end{subfigure}
%     \hspace*{4em} %add desired spacing between images, e. g. ~, \quad, \qquad, \hfill etc. 
%       %(or a blank line to force the subfigure onto a new line)
%     \begin{subfigure}[b]{0.4\textwidth}
%         \includegraphics[width=\textwidth]{bleu_adequacy_new.pdf}
%         \caption{\label{fig:bleu-adequacy}Comparison of BLEU scores to average adequacy scores.}
%     \end{subfigure}
%     \begin{subfigure}[b]{0.4\textwidth}
%         \includegraphics[width=\textwidth]{bert_fluency.pdf}
%         \caption{\label{fig:bert-fluency}Comparison of BERT scores to average fluency scores.}
%     \end{subfigure}
%     \hspace*{4em} %add desired spacing between images, e. g. ~, \quad, \qquad, \hfill etc. 
%       %(or a blank line to force the subfigure onto a new line)
%     \begin{subfigure}[b]{0.4\textwidth}
%         \includegraphics[width=\textwidth]{bert_adequacy.pdf}
%         \caption{\label{fig:bert-adequacy}Comparison of BERT scores to average adequacy scores.}
%     \end{subfigure}
%     %\setlength{\belowcaptionskip}{-.25cm}
%     \caption{Comparison of BLEU and BERT scores to human judgments.
%     \esm{Probably cut these for space; info is in table}}\label{fig:judgments_metrics}
% \end{figure*}

\begin{table*}[htb]
\centering\small
\begin{tabular}{|l|cc|cc|cc|cc|cc|}
\hline \multirow{2}{*}{System} & \multicolumn{2}{c|}{BLEU$_\uparrow$} & \multicolumn{2}{c|}{METEOR$_\uparrow$} & \multicolumn{2}{c|}{TER$_\downarrow$}& \multicolumn{2}{c|}{CHRF++$_\uparrow$}&
\multicolumn{2}{c|}{BERTScore$_\uparrow$}\\
& All & Sub. & All & Sub. & All & Sub. & All & Sub. & All & Sub.\\
\hline
Konstas & \textbf{33.7} & \textbf{38.1} & \textbf{37.4} & \textbf{39.2} & \textbf{48.6} & 45.1 & \textbf{61.0} & \textbf{64.3} & \textbf{94.3} & \textbf{95.0}\\
\hline
Zhu & 31.3 & \textbf{38.1} & 36.2 & 38.7 & 50.8 & \textbf{44.2} & 54.1 & 56.3 & 92.2 & 92.7\\
\hline
Ribeiro & 26.4 & 31.9 & 33.8 & 35.8 & 59.2 & 53.8 & 50.1 & 52.1 & 91.1 & 92.4\\
\hline
Guo & 26.3 & 28.1 & 33.8 & 35.0 & 59.5 & 56.7 & 50.0 & 50.2 & 91.8 & 92.4\\
\hline
Manning & \hphantom{0}9.7 & 10.6 & 28.3 & 28.1 & 68.8 & 67.6 & 47.5 & 48.5 & 89.6 & 89.8 \\
\hline
\end{tabular}
\caption{\label{tab:metrics} Each system's scores on automatic metrics for the full dataset of 1371 sentences (All) and the subset of 100 sentences used in the human evaluation (Sub.).
}
\end{table*}

\subsection{Comparison to Automatic Metrics}

To investigate how well automatic metrics align with human judgments of the relative quality of these systems, we compute BLEU \citep{papineni-etal-2002-bleu}, METEOR \citep{banerjee-lavie-2005-meteor}, TER \citep{Snover2006}, and CHRF++ \citep{popovic-2017-chrf}, and BERTScore \citep{zhang2019bertscore} for each system.\footnote{For reproducibility, details on scripts and parameters used for each metric are given in the supplementary material.} Results are shown in \cref{tab:metrics}. %the relationship between each system's average fluency and adequacy scores to its BLEU score and BERTScore are also visualized in \cref{fig:judgments_metrics}.

All these metrics at least agree with humans that the Konstas and Zhu systems are the best, followed by Ribeiro and Guo, and that Manning is the worst.

Within the top two, humans found Konstas substantially better than Zhu. When using the full data, all automatic metrics agree that Konstas is best, although for all but CHRF++ this is by a small margin. When evaluated only on sentences used in the human evaluation, only METEOR, CHRF++, and BERTScore preserve this ranking; BLEU finds the two essentially tied, while TER finds Zhu slightly better.

For the middle two, humans preferred Ribeiro on fluency but preferred Guo on adequacy. On the full dataset, all the metrics capture that these systems are of very similar overall quality, varying only by a fraction of a point. On the subset of sentences, all metrics except BERTScore prefer Ribeiro, suggesting that these metrics may align more with human judgments of fluency than of adequacy.

Overall, these results show that these metrics essentially capture human rankings of these systems on this dataset, although further research would be needed to more robustly confirm the validity of these metrics for the task.

The results also highlight the limitations of metrics that produce only single scores. While these metrics can only capture that the Ribeiro and Guo systems are similar, our human evaluation found more nuance by identifying criteria on which each one outperforms the other.

% original
% \begin{table}
% \centering\small
% \begin{tabular}{|l|rr|}
% \hline & Fluency & Adequacy \\ \hline
% BLEU$_\uparrow$ & 0.40 & 0.52 \\
% METEOR$_\uparrow$ & 0.41 & 0.57 \\
% TER$_\downarrow$ & -0.33 & -0.43 \\
% CHRF++$_\uparrow$ & 0.32 & 0.47 \\
% BERTScore$_\uparrow$ & \textbf{0.47} & \textbf{0.60} \\ 
% \hline
% \end{tabular}
% \caption{\label{tab:sentence-correlations} Sentence-level correlations of each metric with average human judgments for fluency and adequacy.}
% \end{table}

% transposed version
\begin{table}[htb]
\centering\small
\begin{tabular}{|l|ccccc|}
\hline System & BLEU$_\uparrow$ & METEOR$_\uparrow$ & TER$_\downarrow$ & CHRF++$_\uparrow$ &
BERTScore$_\uparrow$\\
\hline
Fluency & 0.40 & 0.41 & -0.33 & 0.32 & \textbf{0.47} \\ 
Adequacy & 0.52 & 0.57 & -0.43 & 0.47 & \textbf{0.60} \\ 
\hline
\end{tabular}
\caption{\label{tab:sentence-correlations} Sentence-level correlations of each metric with average human judgments.}
\end{table}

Since all these metrics give similar results on system-level rankings, we also calculate each metric's sentence-level correlation with human judgments for adequacy and fluency (each averaged over the two annotator's scores) for more insight into the relative abilities of the metrics to capture human judgments. Results are shown in \cref{tab:sentence-correlations}. We find that each metric correlates more strongly with adequacy than with fluency (perhaps related to the higher IAA for adequacy judgments), and that BERTScore has the strongest correlation with human judgments of both. Our results indicate that BERTScore is the strongest automatic metric for evaluating AMR generation, and that METEOR also appears slightly more reliable than BLEU.

\subsection{Analysis of Adequacy Errors}
\label{ssec:low-adequacy}

To examine what factors contributed to particularly low adequacy scores, we identify sentences for which both annotators gave low ratings. Because, as shown in \cref{fig:annotator-violins}, individual annotators differed in the distribution of ratings they used, we normalized this by annotator: a sentence is counted as low-adequacy if each annotator gave it a rating in the lower 1/3 of their total adequacy ratings. The number of low-scoring sentences by system is given in \cref{tab:low-scores}.

All 139 low-adequacy sentences were marked as having at least one adequacy error by at least one annotator. 46 (33\%) were tagged by both annotators as incomprehensible, 51 (37\%) as missing information, and 25 (18\%) as adding information.

Added information is perhaps the most troubling form of error; AMR generation systems will have severely limited potential for use in practical applications as long as they hallucinate meaning. In one example, a reference to prostitution is inserted:

\textit{REF: A high-security Russian laboratory complex storing anthrax, plague and other deadly bacteria faces loosing electricity for lack of payment to the mosenergo electric utility.}

\textit{RIBEIRO: the russian laboratory complex as a high - security complex will be faced with anthrax , prostitution , and and other killing bacterium losing electricity as it is lack of paying for mosenergo .}

As seen in \cref{tab:avg-scores}, Manning omits and adds information substantially less often than the other systems, but produces incomprehensible sentences far more often. Thus, it is unsurprising that most (73\%) of its low-adequacy sentences are also low-fluency. For Guo, too, a majority (54\%) of low-adequacy sentences are low-fluency, though this is largely due to anonymization and repetition of words, discussed below.

\subsection{Analysis of Fluency Errors}
\label{ssec:low-fluency}

Using the same procedure described above for low adequacy, we also identify sentences for which both annotators gave low fluency ratings. Counts for each system are given in \cref{tab:low-scores}. As expected, no reference sentences are low-fluency.

Of the 116 low-fluency sentences, 50 (43\%) are also marked as incomprehensible by both annotators. The other error types are, unsurprisingly, less related to low fluency than to low adequacy: 23 (20\%) of low-fluency sentences are missing information, and only 6 (5\%) have added information. 

Over half of all low-fluency sentences are from Manning's rule-based system. This is largely because in many cases the system's rules do not allow for the generation of function words that would be expected in a fluent version of the sentence, while the neural systems are more likely to include such words in similar ways to the training data. For example, for the following AMR:\\

\noindent\texttt{(t / thank-01\\
 \indent     :ARG1 (y / you)\\
  \indent    :ARG2 (r / read-01\\
    \indent  \indent  :ARG0 y))}\\
    
\noindent Manning's system gave the disfluent output `\textit{Thank you read .}' while others produced variants of `\textit{thank you for reading .}' or `\textit{thanks for reading .}'

For the neural systems, common sources of low fluency scores included anonymization and repetition of words. Anonymization was a problem primarily for Guo; 9 of Guo's 21 low-fluency sentences contain the token \textit{<unk>} in place of lower-frequency words. For example, for the AMR in \cref{sec:intro}, `annexation' is lost:

\textit{GUO: georgia labels russia 's support for the <unk> act .}

While Konstas uses anonymization less frequently, 2 of the system's 5 low-fluency sentences contain anonymized location names or quantities.

Guo, Ribeiro, and Konstas all have several low-fluency sentences with unhumanlike repetition of words or phrases, for example:\\

\noindent\texttt{(a / and\\
\indent  :op2 (h / happen-02\\
\indent\indent         :ARG1 (l / like-01\\
\indent\indent\indent                 :ARG0 (i / i)\\
\indent\indent\indent                 :ARG1 (d / develop-02
:ARG1 (l2 / lot :mod (l3 / large))))))}\\
\textit{RIBEIRO: and i happen to like a large lot of a lot .}

\section{Conclusion and Future Work}

Our analysis of these systems, and especially of their common errors, points toward directions for researchers developing NLG systems, especially for AMR, to improve their output. We recommend attempting to find solutions to the common issues that led to low scores even from state-of-the-art systems, such as anonymization of infrequent concepts, unnecessary repetition of words, and hallucination.

While this study found that popular automatic metrics were mostly successful in ranking these systems in the same order human annotators did, we also found that the human evaluation was able to identify strengths and weaknesses of systems with more nuance than a single number can convey. We also acknowledge that, given prior work pointing to the inadequacy of metrics such as BLEU for NLG and AMR generation, more research is needed to determine the reliability of these metrics for comparing systems. We suggest that researchers in AMR generation and other NLG tasks continue to supplement automatic metrics with human evaluation as much as possible.

\section*{Acknowledgements}
We thank Austin Blodgett, Lucia Donatelli, Janet Yang Liu, Sean MacAvaney, Siyao Logan Peng, Sasha Slone, and Yilun Zhu for annotating; and Ioannis Konstas, Junhui Li, Jonathan May, Leonardo Ribeiro, and Yan Zhang for sharing their data with us. We also thank everyone who gave us feedback on earlier drafts of this paper, including Emiel van Miltenburg, Gabriel Satanovsky, Amir Zeldes, and anonymous reviewers.

\section*{Appendix A. Survey Details}

The instructions shown to annotators at the start of each section are given below.

Examples chosen for the instructions were items from the pilot where all annotators agreed on a very high or very low score, or marked the applicable error type.

\subsection*{Fluency Instructions}

In this first section you will see several sentences (or utterances) on each page. Please use the slider to indicate how well each one represents fluent English, like you might expect a person who is a native speaker of English to use. Some of these may be sentence fragments rather than complete sentences, but can still be considered fluent utterances.

The left end of the slider represents the worst (least fluent) utterances; here is an example of something that might get a very low rating:\\

\textit{Effective remov 300,000,000 acres land total oil explore market}\\

The right end of the slider represents the best (most fluent) utterances; those that you might expect a fluent English speaker to write. Here is an example of something that might get a very high rating:\\

\textit{in total, more than 300 million onshore acres of federal land have been effectively removed from the market for oil exploration.}\\

The slider gives you the freedom to represent many different levels of fluency between these extremes. Try to give it a similar position for utterances that you think are at the same level of fluency, but since it can be hard to be exact, don't worry too much about very small differences in slider position.

\subsection*{Adequacy Instructions}
In this section you will again see several utterances on each page. In this case, you will also see the AMR that each one should match in meaning. Your tasks is to determine how accurately the sentence expresses the meaning in the AMR.

The left end of the slider represents utterances that do not match the AMR's meaning at all. The example below would receive a very low score, because the sentence contains almost none of the information in the AMR:\\

\texttt{
(p / possible-01 :polarity -\\
\indent      :ARG1 (t / tell-01 \\
\indent\indent            :ARG0 (i / i)\\
\indent\indent            :ARG1 (m / many\\
\indent\indent\indent                  :frequency-of (f / fantasize-01 \\
\indent\indent\indent\indent                        :ARG0 i \\
\indent\indent\indent\indent                        :ARG1 (o / or \\
\indent\indent\indent\indent\indent                              :op1 (p2 / pizza \\
\indent\indent\indent\indent\indent\indent                                    :source (c / company :wiki "Pizza\_Hut"\\
\indent\indent\indent\indent\indent\indent\indent                                          :name (n / name :op1 "Pizza" :op2 "Hut")))\\
\indent\indent\indent\indent\indent                             :op2 (t2 / thing :wiki "Whopper"\\
\indent\indent\indent\indent\indent\indent                                    :name (n2 / name :op1 "Whopper")\\
\indent\indent\indent\indent\indent\indent                                    :ARG0-of (h / have-03\\
 \indent\indent\indent\indent\indent\indent\indent                                         :ARG1 (c2 / cheese))))))\\
\indent\indent            :ARG2 (y / you)))}

    \textit{My tell you}\\
    
    The right end represents utterances that represent the meaning in the AMR completely and accurately, for example, for the same AMR as above:\\
    
    \textit{i can't tell you how many times i would fantasize about a pizza hut pizza or a whopper with cheese.}\\
    
    In addition to rating each sentence with a slider, you will have a space to indicate particular types of errors a sentence may have. Please choose any that you feel applies.
    
    "Cannot understand meaning" is used when the utterance is so nonsensical that you cannot determine its `meaning' in order to compare it to the AMR in a meaningful way. Here's an example that would get this rating:\\
    
    \textit{no one could have pizza from pizza hut has a cheese whopper or fantasize, i told you many.}\\
    
    "Information from the AMR is missing in the utterance" is used when the sentence does not express all the information that is provided in the AMR. The "My tell you" example above would receive this.
    
    "Information in the utterance is not present in the AMR" is used when the sentence expresses information that is not provided in the AMR. The following is an example that would receive this, because the "should" modality expressed in the sentence is not in the AMR:\\
    
\texttt{
(s / say-01\\
\indent      :ARG0 (p / person :wiki "Howard\_Weitzman" :name (n / name :op1 "Howard" :op2 "Weitzman"))\\
\indent       :ARG1 (c / compensate-01\\
\indent \indent             :ARG2 (t / they)\\
\indent \indent             :degree (f / fair))}\\
    
    \textit{howard weitzman , said they should be fairly compensated}\\
    
    The utterances on each page all correspond to the same AMR; the AMR is repeated with each sentence so that you can easily compare it to each one.
    
    You may also wish to consult the AMR guidelines (\url{https://github.com/amrisi/amr-guidelines/blob/master/amr.md}). %\href{https://github.com/amrisi/amr-guidelines/blob/master/amr.md}{AMR guidelines}. % href seems to break coling

\section*{Appendix B. Automatic Metrics Details}
Before computing automatic reference scores, we detokenize references and system outputs using NLTK's TreebankWordDetokenizer \citep{Bird2009} to normalize any tokenization differences.

We compute BLEU with SacreBLEU\footnote{\url{https://github.com/mjpost/sacreBLEU}} \citep{post-2018-call}. We make it case-insensitive (\texttt{-lc}) and otherwise use default parameters.

For METEOR, we use Version 1.5\footnote{\url{https://www.cs.cmu.edu/~alavie/METEOR/}} \citep{denkowski-lavie-2014-meteor} with English normalization (\texttt{-l en -norm}).

For TER, we use Version 0.7.25\footnote{\url{http://www.cs.umd.edu/~snover/tercom/}} with normalization (\texttt{-N}).

For CHRF++\footnote{\url{https://github.com/m-popovic/chrF}}, we use default settings.

For BERTScore, we use version 0.3.1\footnote{\url{https://github.com/Tiiiger/bert_score}}, with default English settings (\texttt{roberta-large\_L17\_no-idf\_version=0.3.1 (hug\_trans=2.4.1)}).

\bibliography{anthology,NLG_eval}
\bibliographystyle{acl_natbib}

\end{document}